\documentclass[journal]{IEEEtran}

\usepackage{graphicx, times, amsmath, amssymb, cite, subfigure, stfloats, subfigure,booktabs, mathtools,bm,threeparttable,multirow}
\usepackage[ruled,norelsize]{algorithm2e}

\makeatletter
\newcommand{\removelatexerror}{\let\@latex@error\@gobble}
\makeatother

\begin{document}

\title{Optimize TSK Fuzzy Systems for Classification Problems: Mini-Batch Gradient Descent with Uniform Regularization and Batch Normalization}
\author{
    Yuqi~Cui, Dongrui~Wu and Jian~Huang
    \thanks{Y.~Cui, D.~Wu and J.~Huang are with the Key Laboratory of the Ministry of Education for Image Processing and Intelligent Control, School of Artificial Intelligence and Automation, Huazhong University of Science and Technology, Wuhan 430074, China. Email: yqcui@hust.edu.cn, drwu@hust.edu.cn, huang\_jan@hust.edu.cn.}
    \thanks{D.~Wu and J.~Huang are the corresponding authors.}
}
\maketitle

\begin{abstract}
Takagi-Sugeno-Kang (TSK) fuzzy systems are flexible and interpretable machine learning models; however, they may not be easily optimized when the data size is large, and/or the data dimensionality is high. This paper proposes a mini-batch gradient descent (MBGD) based algorithm to efficiently and effectively train TSK fuzzy classifiers. It integrates two novel techniques: 1) uniform regularization (UR), which forces the rules to have similar average contributions to the output, and hence to increase the generalization performance of the TSK classifier; and, 2) batch normalization (BN), which extends BN from deep neural networks to TSK fuzzy classifiers to expedite the convergence and improve the generalization performance. Experiments on 12 UCI datasets from various application domains, with varying size and dimensionality, demonstrated that UR and BN are effective individually, and integrating them can further improve the classification performance.
\end{abstract}


\section{Introduction}\label{sec:intro}

Takagi-Sugeno-Kang (TSK) fuzzy systems~\cite{nguyen2019fuzzy} have achieved great success in numerous applications, including both classification and regression problems. Many optimization approaches have been proposed for them.

There are generally three strategies for fine-tuning the TSK fuzzy system parameters after initialization: 1) evolutionary algorithms \cite{shi1999implementation,wu2006genetic}; 2) gradient descent (GD) based algorithms \cite{Wang1992b}; and, 3) GD plus least squares estimation (LSE), represented by the popular adaptive-network-based fuzzy inference system (ANFIS) \cite{jang1993anfis}. However, these approaches may have challenges when the size and/or the dimensionality of the data increase. Evolutionary algorithms need to keep a large population of candidate solutions, and evaluate the fitness of each, which result in high computational cost and heavy memory requirement for big data. Traditional GD needs to compute the gradients from the entire dataset to iteratively update the model parameters, which may be very slow, or even impossible, when the data size is very large. The memory requirement and computational cost of LSE also increase rapidly when the data size and/or dimensionality increase. Additionally, as shown in \cite{drwuGD2019}, ANFIS may result in significant overfitting in regression problems.

Many efforts have been spent to tackling the difficulty in optimizing the TSK fuzzy systems on big and/or high-dimensional data~\cite{jin2000fuzzy, deng2016hierarchical,chung2008minimum}. Dimensionality reduction and/or feature selection are usually used to reduce the number of fuzzy partitions (rules). Traditional dimensionality reduction techniques such as principal component analysis (PCA) has been used for TSK fuzzy system optimization~\cite{nilashi2015multi, lau2013fault}. There are also methods focusing on learning a sparse subspace of the original feature space to reduce the number of antecedents in each rule~\cite{deng2016survey,deng2010enhanced}. Once the number of antecedents is determined, different optimization approaches can be used to tune the TSK fuzzy system on large datasets. For example, Chung \emph{et al.} \cite{chung2008minimum} utilized the equivalence between minimum enclosing ball and the Mamdani-Larsen fuzzy inference system to train the latter using the former. Gacto \emph{et al.} \cite{gacto2014metsk} proposed a multi-objective evolutionary algorithm to optimize TSK fuzzy systems for high-dimensional large-scale regression problems.

Mini-batch gradient descent (MBGD)~\cite{goodfellow2016deep,Ruder2016} based optimization, which is particularly popular in deep learning, can also be a solution to training TSK fuzzy systems on large and high-dimensional datasets. In each iteration, MBGD computes the gradients from a randomly selected small batch of data, instead of the entire dataset~\cite{bottou2010large}. Different batch sizes can be used, according to the trade-off among the available memory, the training speed, and the expected generalization performance. The original MBGD used a constant learning rate to update the model's parameters~\cite{bottou2010large}. Later, Sutskever \emph{et al.}~\cite{sutskever2013importance} found that adding a momentum to MBGD can improve the final training performance. However, it still needs to manually select a learning rate, and the convergence may be very slow at the beginning. Kingma and Ba~\cite{kingma2015adam} proposed the well-known Adam algorithm to automatically rescale the gradients to achieve adaptive and individualized learning rate for each parameter, which leads to faster convergence. However, the generalization performance of Adam may not be as good as the momentum~\cite{wilson2017marginal}; so, Keskar and Socher~\cite{keskar2017improving} also tried to combine the advantages of momentum and Adam to achieve both fast convergence and good generalization. Recently, Luo \emph{et al.}~\cite{luo2019adaptive} also proposed AdaBound to improve Adam. AdaBound uses an adaptive bound for the learning rate of each parameter to force the optimizer to behave like Adam at the beginning and like stochastic GD at the end. Our very recent research \cite{drwuGD2019} has found that TSK fuzzy systems can achieve better performance with AdaBound than Adam for regression problems.

Although MBGD-based optimization has many advantages, it may be easily trapped into a local-minimum, and may face the gradient vanishing problem. Many other techniques have been proposed to complement MBGD for better performance. In 2015, Ioffe and Szegedy~\cite{ioffe2015batch} proposed the well-known batch normalization (BN) approach to accelerate the training of deep neural networks by reducing the internal covariate shift\footnote{Recently some researchers had different opinions on why BN works. For example, Santurkar \emph{et al.}~\cite{santurkar2018does} argued that BN may not reduce the internal covariate shift; instead, it helps improve the Lipschitzness of both the loss and the gradients, and also reduces the dependency on the training hyper-parameters, such as the learning rate and the regularization weights.}. BN normalizes the input distribution of each layer, so it also alleviates the gradient vanishing problem. It has been used almost ubiquitously in deep learning, and many variants \cite{ba2016layer,fan2017revisit,clevert2015fast,wu2018group} have also been proposed.

This paper, following our previous research \cite{drwuGD2019} on MBGD-based optimization of TSK fuzzy systems for regression problems, considers classification problems. We use AdaBound, as in \cite{drwuGD2019}, to adjust the learning rates. Additionally, we propose two novel techniques for training TSK fuzzy systems for classification problems, namely, uniform regularization (UR) and BN. Our main contributions are:
\begin{enumerate}
\item We introduce a novel UR term to the cross-entropy loss function in training TSK fuzzy classifiers, which forces all rules to have similar average firing levels on the entire dataset. Experiments show that UR can improve the generalization performance of TSK fuzzy classifiers.
\item We extend BN from the training of deep neural networks to the training of TSK fuzzy classifiers, and show that it can speed up the convergence in training and improve the generalization performance in testing.
\item We further integrate UR and BN, and show that the combined approach outperforms each individual ones.
\end{enumerate}

The remainder of this paper is organized as follows: Section~\ref{sec:methods} introduces the proposed UR and BN approaches. Section~\ref{sec:res} presents the experimental results to validate the performances of UR and BN. Section~\ref{sec:conclusion} draws conclusions and points out some future research directions.

\section{UR and BN}\label{sec:methods}

This section introduces the details of the TSK fuzzy classifier under consideration, our proposed UR for regularizing the loss function, and BN for more efficient and effective training of the TSK fuzzy classifier. Python implementation of our algorithm can be downloaded at \emph{https://github.com/YuqiCui/TSK\_BN\_UR}.

\subsection{The TSK Fuzzy Classifier}

Let the training dataset be $\mathcal{D} = \{\bm{x}_n, y_n\}_{n=1}^{N}$, in which $\bm{x}_n=[x_{n,1},...,x_{n,D}]^T\in \mathbb{R}^{D \times 1}$ is a $D$-dimensional feature vector, and $y_n \in \{1,2,...,C\}$ the corresponding class label for a $C$-class classification problem.

Suppose the TSK fuzzy classifier has $R$ rules, in the following form:
\begin{align}
	\begin{split}
		\textup{Rule}_r:~&\textup{IF}~x_1~\textup{is}~X_{r, 1}~\textup{and}~ \cdots ~\textup{and}~x_D~\textup{is}~X_{r, D}, \\
		&\textup{THEN}~y_r^1(\bm{x}) = b_{r,0}^1+\sum_{d=1}^{D}b_{r,d}^1\cdot x_d~\textup{and} \cdots \\
&\qquad\qquad \textup{and}~y_r^C(\bm{x}) = b_{r,0}^C+\sum_{d=1}^{D}b_{r,d}^C\cdot x_d
	\end{split} \label{eq:rule}
\end{align}
where $X_{r, d}$ ($r=1,...,R$; $d=1,...,D$) is the membership function (MF) for the $d$-th antecedent in the $r$-th rule, and $b_{r,0}^c$ and $b_{r,d}^c$ ($c=1,...,C$) are the consequent parameters for the $c$-th class.

Different types of MFs can be used in our algorithm, as long as they are differentiable. For simplicity, Gaussian MFs are considered in this paper, and the membership grade of $x_d$ on $X_{r,d}$ is:
\begin{align}
	\mu_{X_{r, d}}(x_d) &= \exp\left(-\frac{(x_d-m_{r,d})^2}{2\sigma_{r,d}^2}\right), \label{eq:MG}
\end{align}
where $m_{r,d}$ and $\sigma_{r,d}$ are the center and the standard deviation of the Gaussian MF, respectively.

The output of the TSK fuzzy classifier for the $c$-th class is:
\begin{align}
y^c(\bm{x})=\frac{\sum_{r=1}^Rf_r(\bm{x})y_r^c(\bm{x})}{\sum_{r=1}^Rf_r(\bm{x})}, \label{eq:yx}
\end{align}
where
\begin{align}
f_r(\bm{x})=\prod_{d=1}^D \mu_{X_{r,d}}(x_d) =\textup{exp}\left(-\sum_{d=1}^D\frac{(x_d-m_{r,d})^2}{2\sigma_{r,d}^2}\right) \label{eq:prod}
\end{align}
is the firing level of Rule~$r$. We can also re-write (\ref{eq:yx}) as:
\begin{align}
y^c(\bm{x})=\sum_{r=1}^R\overline{f}_r(\bm{x})y_r^c(\bm{x}),
\end{align}
where
\begin{align}
\overline{f}_r(\bm{x})=\frac{f_r(\bm{x})}{\sum_{i=1}^R f_i(\bm{x})}
\end{align}
is the normalized firing level of Rule~$r$.

Once the output vector $\bm{y}(\bm{x})=[y^1(\bm{x}),...,y^C(\bm{x})]^T$ is obtained, the input $\bm{x}$ is assigned to the class with the largest $y^c(\bm{x})$.

To optimize the TSK fuzzy classifier, we need to fine-tune the antecedent MF parameters $m_{r,d}$ and $\sigma_{r,d}$, and the consequent parameters $b_{r,0}^c$ and $b_{r,d}^c$, where $r=1,...,R$, $d=1,...,D$, and $c=1,...,C$.

\subsection{Uniform Regularization (UR)}\label{subsec:ur}

\begin{figure}[htbp]\centering
\includegraphics[width=.85\linewidth,clip]{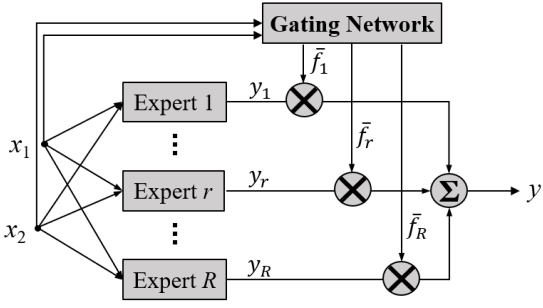}
\caption{Mixture of experts (MoE)~\cite{Jacobs1991}.} \label{fig:MoE}
\end{figure}

Mixture of experts (MoE)~\cite{Jacobs1991}, which is functionally equivalent to TSK fuzzy systems~\cite{Bersini1997,Andersen1998,wu2019functional}, is a popular machine learning algorithm. Its model is shown in Fig.~\ref{fig:MoE}. It trains multiple local experts, each taking care of only a small local region of the input space. For a new input, the gating network determines the activations (weights) of the local experts, and the final output is a weighted average of the local expert outputs.

Although MoE has been used successfully in many applications, it may suffer from the ``rich get richer" effect \cite{shen2019mixture,shazeer2017outrageously}: once an expert is slightly better than others, it is always picked by the gating network, whereas other experts starve and are rarely used. This is bad for the generalization performance of the overall model.

Since MoE and TSK fuzzy systems are functionally equivalent~\cite{wu2019functional}, TSK fuzzy systems may also suffer from the ``rich get richer" effect, i.e., only a few rules are always activated with large firing levels, whereas others have very small firing levels, and hence not adequately tuned in training. A remedy to the ``rich get richer" effect in TSK fuzzy systems is to force the rules to be fired at similar degrees in the input space, so that each rule contributes about equally to the output.

Next, we propose UR to achieve this goal.

UR forces the rules to have similar average firing levels, by minimizing the following loss:
\begin{align}
	\ell_{UR} = \sum_{r=1}^R\left(\frac{1}{N}\sum_{n=1}^N \overline{f}_r(\bm{x}_n) - \tau\right)^2,
\end{align}
where $N$ is the number of training examples, and $\tau$ the expected firing level of each rule, which is set to $1/C$ in this paper (recall that $C$ is the number of classes).

$\ell_{UR}$ can then be added to the original loss function in MBGD-based training of TSK fuzzy classifiers, i.e., for each mini-batch with $N$ training samples,
\begin{align}
	\mathcal{L} = \ell + \alpha \ell_2 + \lambda \sum_{r=1}^R\left(\frac{1}{N}\sum_{n=1}^N\overline{f}_r(\bm{x}_n)-\frac{1}{R}\right)^2, \label{eq:ur}
\end{align}
where $\ell$ is the cross-entropy loss between the estimated class probabilities [obtained by applying \emph{softmax} to $\bm{y}(\bm{x})$] and the true class probabilities, $\ell_2$ the L2 regularization of the rule consequent parameters, and $\alpha$ and $\lambda$ the trade-off parameters.

\subsection{Batch Normalization (BN)}\label{subsec:bn}

BN \cite{ioffe2015batch} is a very powerful technique in optimizing deep neural networks~\cite{he2016deep,zagoruyko2016wide,huang2017densely}. It normalizes the data distribution in each mini-batch to accelerate the training. For a mini-batch $\mathcal{B}=\{\bm{x}_n\}_{n=1}^N$, the output of BN is \cite{ioffe2015batch}:
\begin{align}
\bm{x}'_n = BN(\bm{x}_n) = \gamma \frac{\bm{x}_n-\bm{m}_\mathcal{B}}{\sqrt{\bm{\sigma}_\mathcal{B}^2+\epsilon}}+\beta, \label{eq:bn}
\end{align}
where $\bm{m}_\mathcal{B}$ and $\bm{\sigma}_\mathcal{B}$ are the mean and the standard deviation of the samples in the mini-batch, respectively, $\gamma$ and $\beta$ are parameters to be learned during training, and $\epsilon$ is usually set to $1e-8$ to avoid being divided by zero. During training, exponential weighted averages of $\bm{m}_\mathcal{B}$ and $\bm{\sigma}_\mathcal{B}$ are recorded so that they can be used in the test phase.

\begin{figure}[htpb] \centering
\includegraphics[width=.85\linewidth,clip]{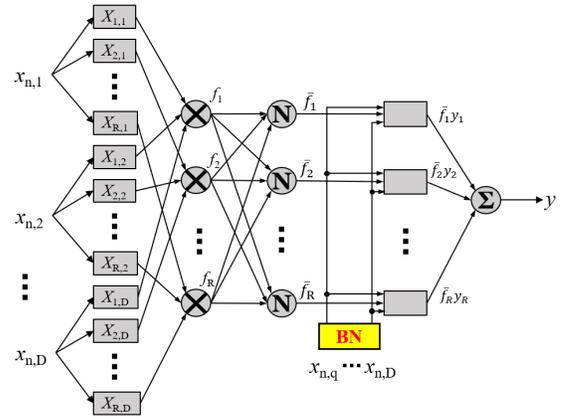}
	\caption{BN in training a TSK fuzzy classifier. All rule consequents share the same BN layer.}
	\label{fig:bn}
\end{figure}

Since TSK fuzzy systems and neural networks share lots of similarity \cite{wu2019functional}, we can extend BN to the optimization of TSK fuzzy classifiers, as shown in Fig.~\ref{fig:bn}. In the training phase, we first compute the firing level of each rule using the unmodified inputs, as in traditional TSK fuzzy systems. Then, we use BN to normalize the inputs, according to their mean and standard deviation in the current mini-batch. The normalized inputs are then used to compute the rule consequents. The final output is a weighted average of the rule consequents, the weights being the corresponding rule firing levels.

At the testing phase, the BN operation can be merged into the consequent layer. Assume that after training, we obtain a BN layer with learned $\bm{m}=(m_1,...,m_D)^T$, $\bm{\sigma}=(\sigma_1,...,\sigma_D)^T$, $\gamma$ and $\beta$. Then, the output $y_r$ of the $r$-th rule with BN is:
\begin{align}
	y_r(BN(\bm{x}_n)) &= b_{r,0}+ \gamma\sum_{d=1}^D b_{r,d}\frac{x_{n,d}-m_d}{\sqrt{\sigma_d^2+\epsilon}} + \beta D,
\end{align}
which can be re-written as:
\begin{align}
	y_r(BN(\bm{x}_n)) = b'_{r,0}+\sum_{d=1}^D b'_{r,d}x_{n,d}, \label{eq:bnrew}
\end{align}
where
\begin{align}
b'_{r,0} &= b_{r,0}+\beta D
-\gamma\sum_{d=1}^D\frac{m_db_{r,d}}{\sqrt{\sigma_d^2+\epsilon}},\\
b'_{r,d}& = \gamma \frac{b_{r,d}}{\sqrt{\sigma_d^2+\epsilon}}.
\end{align}
By doing this, the original architecture of the TSK fuzzy classifier is kept unchanged.

\begin{figure}[htpb] \centering
	\subfigure[]{\includegraphics[width=0.9\columnwidth,clip]{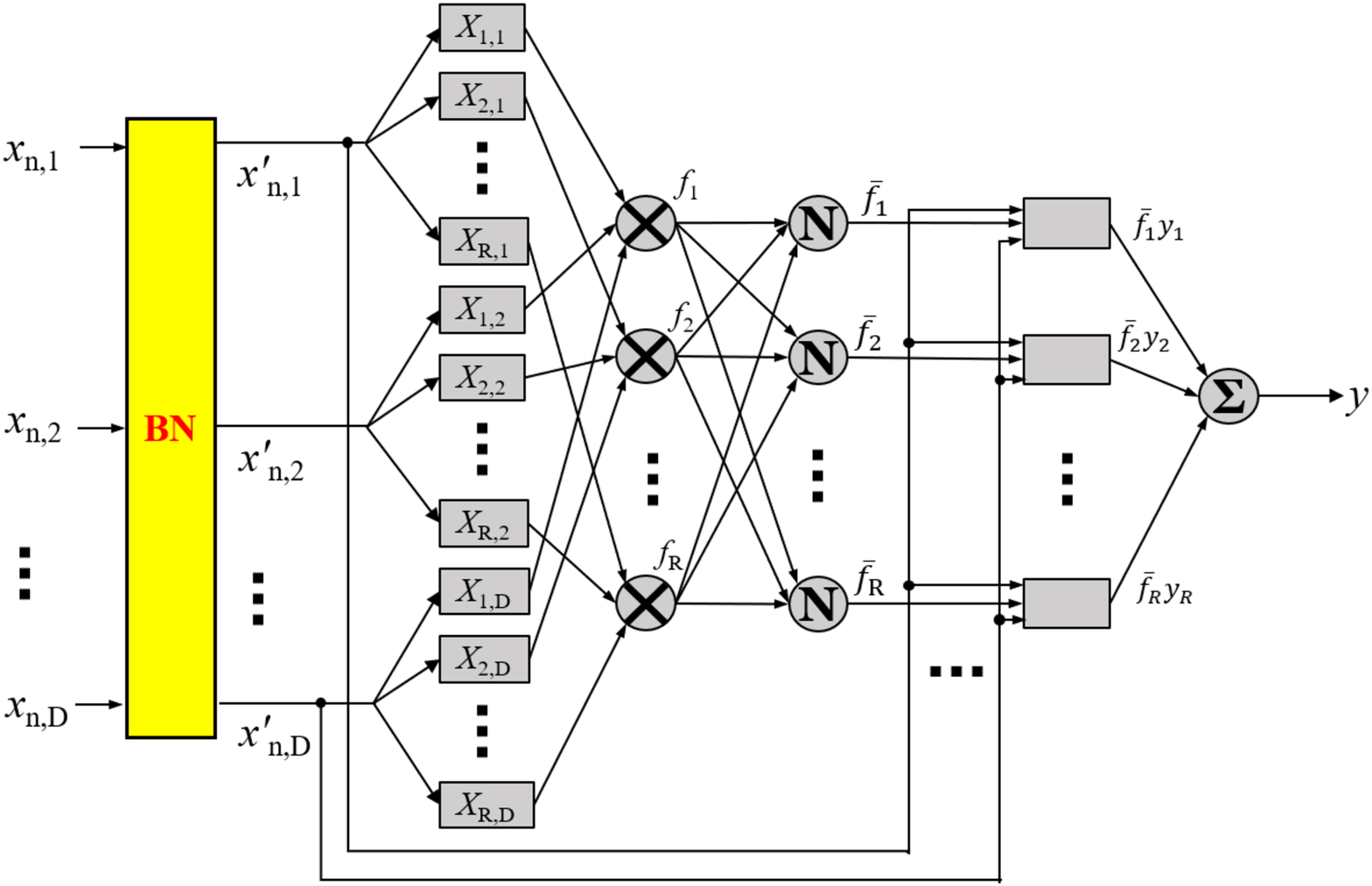}\label{fig:bnv2}}
	\subfigure[]{\includegraphics[width=0.9\columnwidth,clip]{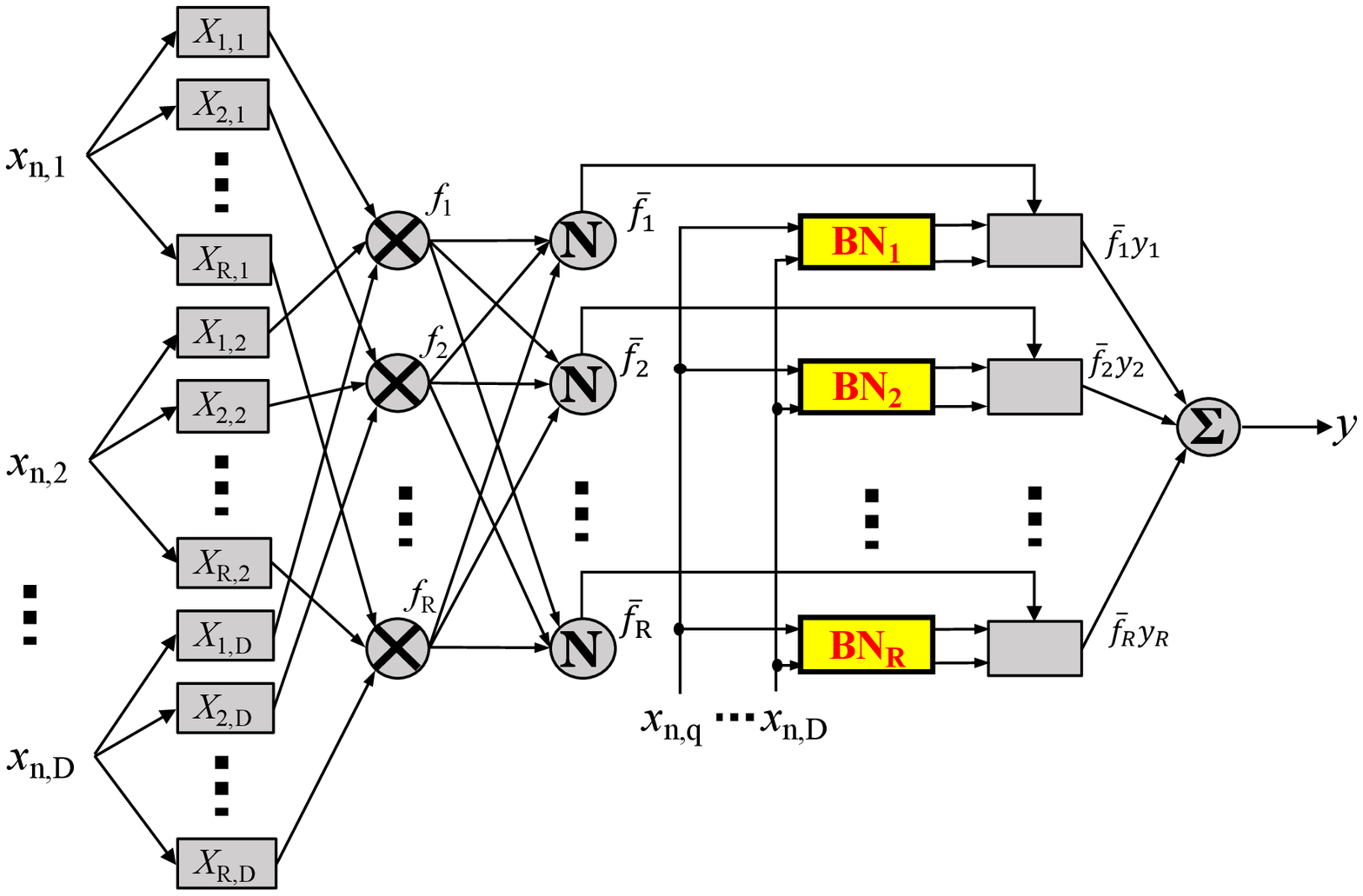}\label{fig:bnv3}}
	\caption{(a) TSK fuzzy system with global BN (\texttt{TSK-MBGD-UR-GBN}); and, (b) TSK fuzzy system with rule-specific BN (\texttt{TSK-MBGD-UR-RBN}).} 	\label{fig:bnv}
\end{figure}

We also tested two variants of BN, as shown in Fig.~\ref{fig:bnv}. The TSK with global BN (\texttt{TSK-MBGD-UR-GBN}) approach in Fig.~\ref{fig:bnv2} uses the BN normalized inputs in both antecedents and consequents to compute the final output. In this case, the output of \texttt{TSK-MBGD-UR-GBN} for Class~$c$ is:
\begin{align}
	y^c(\bm{x})=\sum_{r=1}^R\overline{f}_r(BN(\bm{x}))y_r^c(BN(\bm{x})).
\end{align}
The TSK with rule-specific BN (\texttt{TSK-MBGD-UR-RBN}) approach in Fig.~\ref{fig:bnv3} uses the raw inputs to compute the antecedents, and rule-specific BN to compute each consequent individually. The output of \texttt{TSK-MBGD-UR-RBN} for Class~$c$ is:
\begin{align}
	y^c(\bm{x})=\sum_{r=1}^R\overline{f}_r(\bm{x})y_r^c(BN_r(\bm{x})),
\end{align}
where $BN_r$ represents the BN operation for the $r$-th rule.

\texttt{TSK-MBGD-UR-GBN} has the same computational cost as \texttt{TSK-MBGD-UR-BN}, but \texttt{TSK-MBGD-UR-RBN} has $R$ times more BN parameters, and hence higher computational cost. Both of them can be re-expressed in the original TSK architecture. We also evaluate their performances in Section~\ref{subsec:bnres}.

\section{Experiments and Results}\label{sec:res}

This section validates the performances of our proposed UR and BN on multiple datasets from various application domains, with varying size and feature dimensionality.

\subsection{Datasets}

We evaluated our proposed algorithms on 12 classification datasets from the UCI Machine Learning Repository\footnote{http://archive.ics.uci.edu/ml/index.php}. Their characteristics are summarized in Table~\ref{tab:data}. For each dataset, we randomly selected 70\% samples as the training set and the remaining 30\% as the test set for 30 times to get 30 different data splits. We ran each algorithm on these 30 data splits and report the average performance.

\begin{table}[htpb]
\begin{threeparttable} \setlength{\tabcolsep}{1.5mm}{
\caption{Summary of the 12 datasets.} \label{tab:data}
\begin{tabular}{ccccc} \hline
Index & Dataset                     & No. of Samples & No. of Features & No. of Classes \\ \hline
1 & Vehicle\tnote{1}                     & 846            & 18              & 4              \\
2 & Biodeg\tnote{2}                      & 1,055           & 41              & 2              \\
3 & DRD\tnote{3} & 1151           & 19              & 2              \\
4 & Yeast\tnote{4}                       & 1,484           & 8               & 10             \\
5 & Steel\tnote{5}                       & 1,941           & 27              & 7              \\
6 & IS\tnote{6}           & 2,310           & 19              & 7              \\
7 & Abalone\tnote{7}                     & 4,177           & 10              & 3              \\
8 & Waveform21\tnote{8}                  & 5,000           & 21              & 3              \\
9 & Page-blocks\tnote{9}                & 5,473           & 10              & 5              \\
10 & Satellite\tnote{10}                   & 6,435           & 36              & 6              \\
11 & Clave\tnote{11}                       & 10,798          & 16              & 4              \\
12 & MAGIC\tnote{12}                       & 19,020          & 10              & 2              \\ \hline
\end{tabular}}
\begin{tablenotes}
\item[1] https://archive.ics.uci.edu/ml/datasets/Statlog+\%28Vehicle+Silhouettes\%29
\item[2] https://archive.ics.uci.edu/ml/datasets/QSAR+biodegradation
\item[3] https://archive.ics.uci.edu/ml/datasets/Diabetic+Retinopathy+Debrecen +Data+Set
\item[4] https://archive.ics.uci.edu/ml/datasets/Yeast
\item[5] https://archive.ics.uci.edu/ml/datasets/Steel+Plates+Faults
\item[6] https://archive.ics.uci.edu/ml/datasets/Image+Segmentation
\item[7] https://archive.ics.uci.edu/ml/datasets/Abalone
\item[8] https://archive.ics.uci.edu/ml/datasets/Waveform+Database+Generator +(Version+1)
\item[9] https://archive.ics.uci.edu/ml/datasets/Page+Blocks+Classification
\item[10] https://archive.ics.uci.edu/ml/datasets/Statlog+(Landsat+Satellite)
\item[11] https://archive.ics.uci.edu/ml/datasets/Firm-Teacher\_Clave-Direction\_Classification
\item[12] https://archive.ics.uci.edu/ml/datasets/MAGIC+Gamma+Telescope
\end{tablenotes}
\end{threeparttable}
\end{table}

Some datasets contain both numerical features and categorical features. The categorical features were converted into numerical ones by one-hot coding. We $z$-normalized each feature using the mean and standard deviation computed from the training set.

\subsection{Algorithms}

We compared nine algorithms to validate our proposed approaches. Among them, four were tree based approaches (\texttt{DT}, \texttt{RF}, \texttt{PART}, and \texttt{JRip}), one was a TSK fuzzy system optimized by a traditional approach (\texttt{TSK-FCM-LSE}), and the remaining four were TSK fuzzy systems optimized by MBGD based approaches (\texttt{TSK-MBGD}, \texttt{TSK-MBGD-BN}, \texttt{TSK-MBGD-UR}, \texttt{TSK-MBGD-UR-BN}).

The details of these nine algorithms are as follows:
\begin{enumerate}
\item \texttt{DT}: Decision tree implemented in scikit-learn\footnote{https://scikit-learn.org/stable/modules/generated/sklearn.tree.DecisionTree Classifier.html} in Python. We used 5-fold cross-validation to select the maximum depth of the tree from $\{3, 4, 5, 6, 7\}$ on the training set. Other parameters were set by default.
\item \texttt{RF}: Random forest implemented in scikit-learn\footnote{https://scikit-learn.org/stable/modules/generated/sklearn.ensemble.Random ForestClassifier.html} in Python. We set the number of trees to 20 and used 5-fold cross-validation to select the maximum depth of the trees from $\{3, 4, 5, 6, 7\}$ on the training set. Other parameters were set by default.
\item \texttt{PART}~\cite{frank1998generating}: The PART (partial decision tree) classifier implemented in RWeka\footnote{https://cran.r-project.org/web/packages/RWeka/index.html}. All parameters were set by default.
\item \texttt{JRip}~\cite{cohen1995repeated}: The RIPPER (Repeated Incremental Pruning to Produce Error Reduction) classifier implemented in RWeka. All parameters were set by default.
\item \texttt{TSK-FCM-LSE}~\cite{jang1997neuro}: We used fuzzy $c$-means (FCM) clustering to estimate the antecedent parameters, and LSE with L2 regularization to estimate the consequent parameters.
\item \texttt{TSK-MBGD}: We used MBGD and AdaBound~\cite{luo2019adaptive} to optimize both the antecedent and the consequent parameters.
\item \texttt{TSK-MBGD-UR}: We used MBGD, AdaBound and UR (Section~\ref{subsec:ur}) to optimize both the antecedent and the consequent parameters. The UR weight $\lambda$ in (\ref{eq:ur}) was selected from $\{0.1, 1, 10, 20, 50\}$ by cross-validation on the training set.
\item \texttt{TSK-MBGD-BN}: We used MBGD, AdaBound and BN (Section~\ref{subsec:bn}) to optimize both the antecedent and the consequent parameters.
\item \texttt{TSK-MBGD-UR-BN}: We used MBGD, AdaBound, BN and UR to optimize both the antecedent and the consequent parameters. The UR weight $\lambda$ in (\ref{eq:ur}) was selected from $\{0.1, 1, 10, 20, 50\}$ by cross-validation on the training set.
\end{enumerate}

For \texttt{TSK-FCM-LSE}, \texttt{TSK-MBGD}, \texttt{TSK-MBGD-BN}, \texttt{TSK-MBGD-UR} and \texttt{TSK-MBGD-UR-BN}, we set the L2 regularization weight $\alpha=0.05$, and the number of rules $R=20$. For \texttt{TSK-MBGD}, \texttt{TSK-MBGD-BN}, \texttt{TSK-MBGD-UR} and \texttt{TSK-MBGD-UR-BN}, we set the learning rate of AdaBound to 0.01, following our previous work \cite{drwuGD2019}. In order to make use of all data in the training set and to reduce overfitting simultaneously, we randomly sampled 20\% data from the training set and trained the TSK model with early stopping five times. The maximum epoch number was 2,000, and the patience of early stopping 40. We recorded the number of epochs at stopping in each run, and trained the final model with the average stopping epoch number on the entire training set.

$k$-mean clustering was used in the MBGD-based algorithms (\texttt{TSK-MBGD}, \texttt{TSK-MBGD-BN}, \texttt{TSK-MBGD-UR}, and \texttt{TSK-MBGD-UR-BN}) to initialize the antecedent parameters. We performed $k$-means clustering on the training set, where $k$ equaled $R$, the number of rules. We then initialized the rule centers to the cluster centers, and randomly initialized the standard deviation $\sigma_{r,d}$ from a Gaussian distribution $\mathcal{N}(1, 0.2)$. For the consequent parameters, we set the initial bias of each rule to zero, and the attribute weight $b_{r,d}$ ($r=1, ..., R$; $d=1, ..., D$) randomly from a uniform distribution $U(-1, 1)$.

\subsection{Performance Measures}

The raw classification accuracy (RCA), which is the total number of correctly classified test samples divided by the total number of test samples, was used as our primary performance measure.

Since some datasets have significant class imbalance, in addition to the RCA, we also computed the balanced classification accuracy (BCA), which is the mean of the per-class RCAs, as our second performance measure.

\subsection{Experimental Results}

The average test RCAs and BCAs are shown in Tables~\ref{tab:acc} and \ref{tab:bca}, respectively. The largest value (best performance) on each dataset is marked in bold. To facilitate the comparison, we also show the ranks of the RCAs and BCAs in Tables~\ref{tab:accrank} and \ref{tab:bcarank}, respectively.

\begin{figure}
	\subfigure[]{\includegraphics[width=0.9\columnwidth,clip]{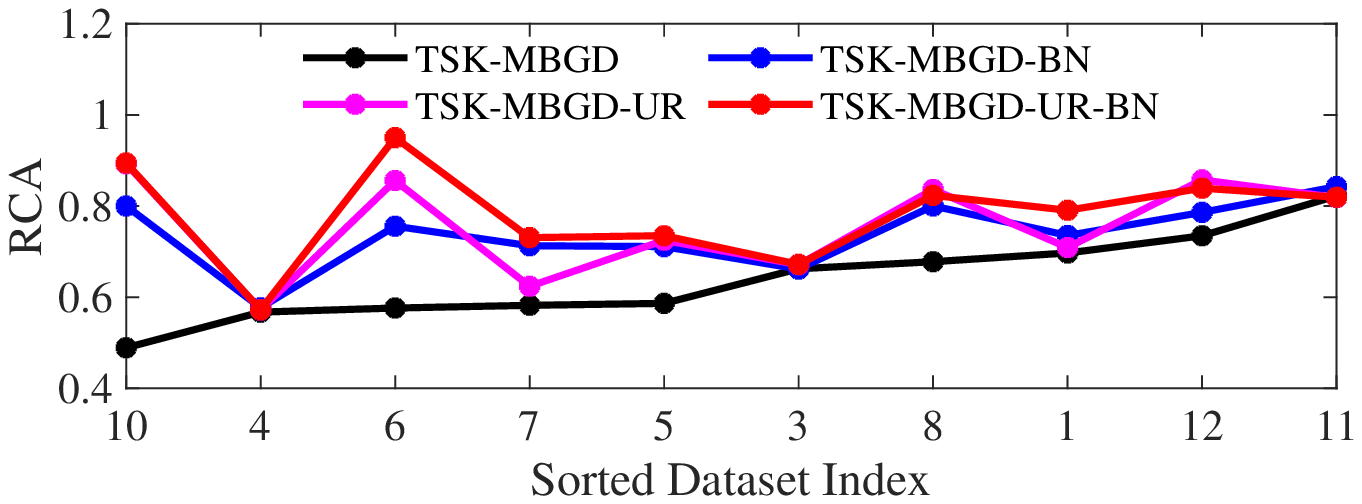}}
	\subfigure[]{\includegraphics[width=0.9\columnwidth,clip]{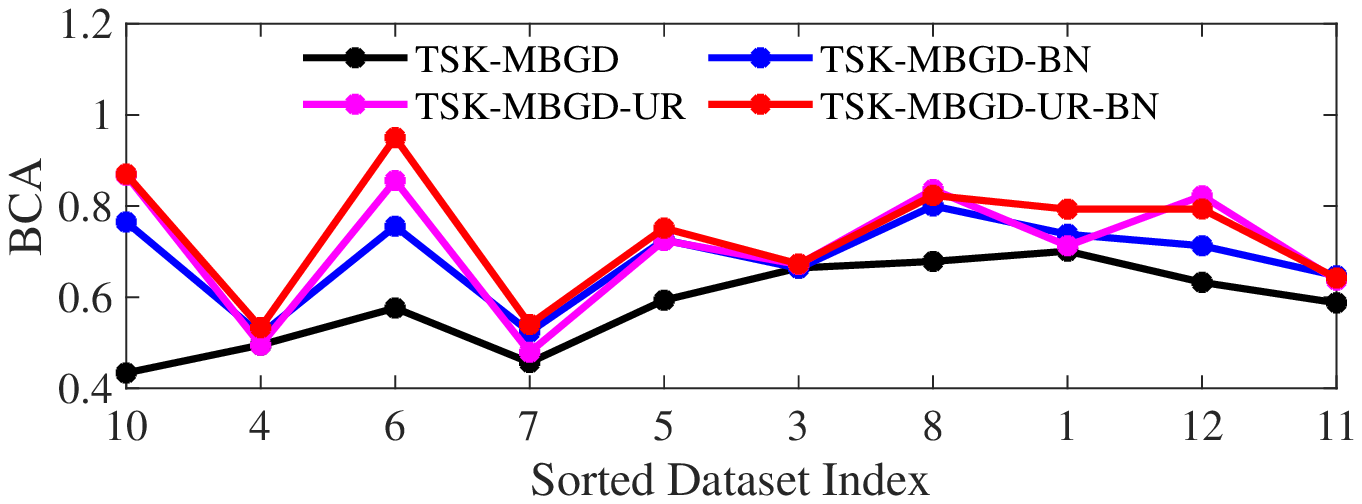}}
	\caption{(a) RCAs and (b) BCAs of the four MBGD-based TSK fuzzy classifiers on the 12 datasets. Datasets were sorted according to the RCAs of the \texttt{TSK-MBGD} model. The indices along the horizontal axis denote the dataset indices in Table~\ref{tab:data}.} \label{fig:improv}
\end{figure}

The following observations can be made from the above four tables:
\begin{enumerate}
	\item \emph{Generally, UR improved both RCA and BCA.} Comparing \texttt{TSK-MBGD} with \texttt{TSK-MBGD-UR}, and \texttt{TSK-MBGD-BN} with \texttt{TSK-MBGD-UR-BN}, we can conclude that generally UR improved the classification performance, regardless of whether BN was used or not. The average ranks in the last row of Tables~\ref{tab:accrank} and \ref{tab:bcarank} demonstrate this more clearly: the average rank of \texttt{TSK-MBGD-UR} (\texttt{TSK-MBGD-UR-BN}) was smaller than that of \texttt{TSK-MBGD} (\texttt{TSK-MBGD-BN}).

	\item \emph{Generally, BN improved both RCA and BCA.} Comparing \texttt{TSK-MBGD} with \texttt{TSK-MBGD-BN}, and \texttt{TSK-MBGD-UR} with \texttt{TSK-MBGD-UR-BN}, we can conclude that generally BN improved the classification performance, regardless of whether UR was used or not. The average ranks in the last row of Tables~\ref{tab:accrank} and \ref{tab:bcarank} demonstrate this more clearly: the average rank of \texttt{TSK-MBGD-BN} (\texttt{TSK-MBGD-UR-BN}) was smaller than that of \texttt{TSK-MBGD} (\texttt{TSK-MBGD-UR}).

	\item \emph{Generally, integrating BN and UR achieved further RCA and BCA improvements.} Comparing \texttt{TSK-MBGD-UR-BN} with \texttt{TSK-MBGD}, \texttt{TSK-MBGD-UR} and \texttt{TSK-MBGD-BN}, we can conclude that \texttt{TSK-MBGD-UR-BN} almost always performed the best on both RCA and BCA, as shown in Fig.~\ref{fig:improv}. This indicated that BN and UR are somehow complementary, and hence integrating them may achieve better performance than using each one alone.

	\item \emph{Overall, \texttt{TSK-MBGD-UR-BN} achieved the best performance among the nine algorithms.}  The last row of Table~\ref{tab:bcarank} shows that \texttt{TSK-MBGD-UR-BN} achieved the best average BCA performance, and the last row of Table~\ref{tab:accrank} shows that \texttt{TSK-MBGD-UR-BN} achieved the second best average RCA performance. Interestingly, \texttt{RF} had the best average rank on RCA, but only ranked the fifth on BCA, suggesting that \texttt{RF} may tend to overlook the minority classes. On the contrary, \texttt{TSK-MBGD-UR-BN} performed well on both RCA and BCA.
\end{enumerate}

\begin{table*}[htpb]\centering \setlength{\tabcolsep}{1.6mm}
\caption{Average RCAs of the nine algorithms on the 12 datasets.} \label{tab:acc}
\begin{tabular}{cccccccccc}\hline
Dataset                     & \texttt{CART}   & \texttt{RF}              & \texttt{JRip}   & \texttt{PART}            & \texttt{TSK-FCM-LSE}             & \texttt{TSK-MBGD}    & \texttt{TSK-MBGD-BN}          & \texttt{TSK-MBGD-UR} & \texttt{TSK-MBGD-UR-BN}       \\ \hline
Vehicle                     & 0.6907 & 0.7407          & 0.6892 & 0.7110          & 0.7411          & 0.6970 & 0.7354          & 0.7089 & \textbf{0.7907} \\
Biodeg                      & 0.8202 & 0.8572          & 0.8222 & 0.8362          & 0.8377          & 0.8523 & 0.8531          & 0.8539 & \textbf{0.8609} \\
DRD & 0.6283 & 0.6589          & 0.6240 & 0.6364          & \textbf{0.6824} & 0.6623 & 0.6618          & 0.6713 & 0.6720          \\
Yeast                       & 0.5564 & \textbf{0.5963} & 0.5731 & 0.5340          & 0.5851          & 0.5673 & 0.5770          & 0.5722 & 0.5725          \\
Steel                       & 0.7017 & 0.7328          & 0.7135 & 0.7120          & 0.6527          & 0.5864 & 0.7110          & 0.7248 & \textbf{0.7350} \\
IS           & 0.9320 & 0.9529          & 0.9481 & \textbf{0.9608} & 0.9571          & 0.5762 & 0.7557          & 0.8559 & 0.9501          \\
Abalone                     & 0.7170 & 0.7314          & 0.7254 & 0.7104          & \textbf{0.7323} & 0.5821 & 0.7129          & 0.6238 & 0.7306          \\
Waveform21                  & 0.7641 & 0.8369          & 0.7908 & 0.7843          & \textbf{0.8647} & 0.6779 & 0.8002          & 0.8363 & 0.8234          \\
Page-blocks                 & 0.9651 & \textbf{0.9688} & 0.9681 & 0.9677          & 0.9499          & 0.9375 & 0.9419          & 0.9515 & 0.9580          \\
Satellite                   & 0.8524 & 0.8863          & 0.8587 & 0.8592          & 0.8864          & 0.4890 & 0.8001          & 0.8929 & \textbf{0.8943} \\
Clave                       & 0.7103 & 0.7600          & 0.7344 & 0.7779          & 0.7690          & 0.8223 & \textbf{0.8427} & 0.8187 & 0.8192          \\
MAGIC                       & 0.8427 & 0.8531          & 0.8455 & 0.8488          & 0.8319          & 0.7347 & 0.7861          & 0.8574 & \textbf{0.8392} \\ \hline
Average                        & 0.7651 & 0.7979          & 0.7744 & 0.7782          & 0.7909          & 0.6821 & 0.7648          & 0.7806 & \textbf{0.8038} \\\hline
\end{tabular}
\end{table*}

\begin{table*}[htpb]\centering \setlength{\tabcolsep}{1.6mm}
\caption{Average BCAs of the nine algorithms on the 12 datasets.} \label{tab:bca}
\begin{tabular}{cccccccccc}
\hline
Dataset                     & \texttt{CART}   & \texttt{RF}              & \texttt{JRip}   & \texttt{PART}            & \texttt{TSK-FCM-LSE}             & \texttt{TSK-MBGD}    & \texttt{TSK-MBGD-BN}          & \texttt{TSK-MBGD-UR} & \texttt{TSK-MBGD-UR-BN}       \\ \hline
Vehicle           & 0.6936 & 0.744  & 0.6939          & 0.7131          & 0.7443          & 0.7010 & 0.7380 & 0.7127          & \textbf{0.7930} \\
Biodeg            & 0.7973 & 0.8306 & 0.7899          & 0.8122          & 0.8205          & 0.8368 & 0.8318 & 0.8390          & \textbf{0.8439} \\
DRD               & 0.634  & 0.6624 & 0.6227          & 0.6422          & \textbf{0.6845} & 0.6642 & 0.6634 & 0.6717          & 0.6729          \\
Yeast             & 0.3998 & 0.4867 & 0.5203          & 0.4889          & 0.5102          & 0.4951 & 0.5184 & 0.4946          & \textbf{0.5332} \\
Steel             & 0.7005 & 0.6937 & 0.7129          & 0.7267          & 0.6319          & 0.5933 & 0.7258 & 0.7245          & \textbf{0.7515} \\
IS & 0.932  & 0.9529 & 0.9481          & \textbf{0.9607} & 0.9571          & 0.5762 & 0.7557 & 0.8559          & 0.9501          \\
Abalone           & 0.5319 & 0.5362 & 0.5371          & 0.5280          & 0.5402          & 0.4567 & 0.5236 & 0.4791          & \textbf{0.5402} \\
Waveform21        & 0.7637 & 0.8365 & 0.7905          & 0.7844          & \textbf{0.8645} & 0.6784 & 0.8003 & 0.8362          & 0.8233          \\
Page-blocks       & 0.7986 & 0.7385 & \textbf{0.8192} & 0.8162          & 0.6003          & 0.5129 & 0.5609 & 0.6033          & 0.671           \\
Satellite         & 0.8204 & 0.8480 & 0.8308          & 0.834           & 0.8558          & 0.4337 & 0.7651 & 0.8679          & \textbf{0.8700} \\
Clave             & 0.4701 & 0.4878 & 0.4985          & \textbf{0.6507} & 0.4825          & 0.5876 & 0.6468 & 0.6374          & 0.6421          \\
MAGIC             & 0.8058 & 0.8108 & 0.8052          & 0.8135          & 0.7886          & 0.6325 & 0.7128 & \textbf{0.8225} & 0.7934          \\ \hline
Average               & 0.6956 & 0.7190 & 0.714           & 0.7309          & 0.7067          & 0.5974 & 0.6869 & 0.7120          & \textbf{0.7404} \\ \hline
\end{tabular}
\end{table*}

\begin{table*}[htpb]\centering
\caption{RCA ranks of the nine algorithms on the 12 datasets.}\label{tab:accrank}
\begin{tabular}{cccccccccc}\hline
Dataset                     & \texttt{CART}   & \texttt{RF}              & \texttt{JRip}   & \texttt{PART}            & \texttt{TSK-FCM-LSE}             & \texttt{TSK-MBGD}    & \texttt{TSK-MBGD-BN}          & \texttt{TSK-MBGD-UR} & \texttt{TSK-MBGD-UR-BN}       \\ \hline
Vehicle                     & 8    & 3   & 9    & 5    & 2   & 7   & 4      & 6      & 1         \\
Biodeg                      & 9    & 2   & 8    & 7    & 6   & 5   & 4      & 3      & 1         \\
DRD & 8    & 6   & 9    & 7    & 1   & 4   & 5      & 3      & 2         \\
Yeast                       & 8    & 1   & 4    & 9    & 2   & 7   & 3      & 6      & 5         \\
Steel                       & 7    & 2   & 4    & 5    & 8   & 9   & 6      & 3      & 1         \\
IS           & 6    & 3   & 5    & 1    & 2   & 9   & 8      & 7      & 4         \\
Abalone                     & 5    & 2   & 4    & 7    & 1   & 9   & 6      & 8      & 3         \\
Waveform21                  & 8    & 2   & 6    & 7    & 1   & 9   & 5      & 3      & 4         \\
Page-blocks                 & 4    & 1   & 2    & 3    & 7   & 9   & 8      & 6      & 5         \\
Satellite                   & 7    & 4   & 6    & 5    & 3   & 9   & 8      & 2      & 1         \\
Clave                       & 9    & 7   & 8    & 5    & 6   & 2   & 1      & 4      & 3         \\
MAGIC                       & 5    & 2   & 4    & 3    & 7   & 9   & 8      & 1      & 6         \\\hline
Average                        & 7.0  & 2.9 & 5.8  & 5.3  & 3.8 & 7.3 & 5.5    & 4.3    & 3.0      \\\hline
\end{tabular}
\end{table*}

\begin{table*}[htbp]\centering
\caption{BCA ranks of the nine algorithms on the 12 datasets.}\label{tab:bcarank}
\begin{tabular}{cccccccccc}
\hline
Dataset                     & \texttt{CART}   & \texttt{RF}              & \texttt{JRip}   & \texttt{PART}            & \texttt{TSK-FCM-LSE}             & \texttt{TSK-MBGD}    & \texttt{TSK-MBGD-BN}          & \texttt{TSK-MBGD-UR} & \texttt{TSK-MBGD-UR-BN}       \\ \hline
Vehicle                     & 9    & 3   & 8    & 5    & 2   & 7   & 4      & 6      & 1         \\
Biodeg                      & 8    & 5   & 9    & 7    & 6   & 3   & 4      & 2      & 1         \\
DRD & 8    & 6   & 9    & 7    & 1   & 4   & 5      & 3      & 2         \\
Yeast                       & 9    & 8   & 2    & 7    & 4   & 5   & 3      & 6      & 1         \\
Steel                       & 6    & 7   & 5    & 2    & 8   & 9   & 3      & 4      & 1         \\
IS           & 6    & 3   & 5    & 1    & 2   & 9   & 8      & 7      & 4         \\
Abalone                     & 5    & 4   & 3    & 6    & 1   & 9   & 7      & 8      & 2         \\
Waveform21                  & 8    & 2   & 6    & 7    & 1   & 9   & 5      & 3      & 4         \\
Page-blocks                 & 3    & 4   & 1    & 2    & 7   & 9   & 8      & 6      & 5         \\
Satellite                   & 7    & 4   & 6    & 5    & 3   & 9   & 8      & 2      & 1         \\
Clave                       & 9    & 7   & 6    & 1    & 8   & 5   & 2      & 4      & 3         \\
MAGIC                       & 4    & 3   & 5    & 2    & 7   & 9   & 8      & 1      & 6         \\ \hline
Average                        & 6.8  & 4.7 & 5.4  & 4.3  & 4.2 & 7.3 & 5.4    & 4.3    & 2.6       \\ \hline
\end{tabular}
\end{table*}

\subsection{Statistical Analysis}

To further evaluate the performance improvement of our proposed \texttt{TSK-MBGD-UR-BN} over others, we also performed non-parametric multiple comparison tests on the RCAs and BCAs using Dunn's procedure~\cite{dunn1964multiple}, with a $p$-value correction using the False Discovery Rate method~\cite{benjamini1995controlling}. The results are shown in Table~\ref{tab:pvalue}, where the statistically significant ones are marked in bold.

Table~\ref{tab:pvalue} demonstrates that our proposed BN and UR can significantly improve the generalization performance of the traditional MBGD optimization for TSK fuzzy classifiers. \texttt{TSK-MBGD-UR-BN} statistically significantly outperformed \texttt{CART}, \texttt{JRip}, \texttt{PART}, \texttt{TSK-MBGD} and \texttt{TSK-MBGD-BN} on RCA, and also statistically significantly outperformed \texttt{CART}, \texttt{JRip}, \texttt{TSK-FCM-LSE}, \texttt{TSK-MBGD} and \texttt{TSK-MBGD-BN} on BCA. Although the performance improvement of \texttt{TSK-MBGD-UR-BN} over \texttt{RF} and \texttt{TSK-MBGD-UR} were not statistically significant, they were quite close to the threshold, especially for the BCA.

\begin{table*}[htpb]\centering
\caption{$p$-values of non-parametric multiple comparisons on the RCAs and BCAs.} \label{tab:pvalue}
\begin{tabular}{cccccccccc}\hline
                           & Metric & \texttt{CART}            & \texttt{RF}              & \texttt{JRip}            & \texttt{PART}            & \texttt{TSK-FCM-LSE}         & \texttt{TSK-MBGD}             & \texttt{TSK-MBGD-BN}          & \texttt{TSK-MBGD-UR} \\\hline
\multirow{2}{*}{\texttt{TSK-MBGD-BN}}    & RCA    & \textbf{0.0097} & 0.0628 & 0.1368          & 0.3239          & 0.2723          & \textbf{0.0000} & -               & -      \\
                           & BCA    & 0.1547          & 0.1627 & 0.4508          & 0.0981          & 0.4518          & \textbf{0.0000} & -               & -      \\\hline
\multirow{2}{*}{\texttt{TSK-MBGD-UR}}    & RCA    & \textbf{0.0001} & 0.3740 & \textbf{0.0090} & 0.0460          & 0.2452          & \textbf{0.0000} & 0.1146          & -      \\
                           & BCA    & \textbf{0.0036} & 0.2900 & 0.0912          & 0.4420          & 0.0921          & \textbf{0.0000} & 0.0731          & -      \\\hline
\multirow{2}{*}{\texttt{TSK-MBGD-UR-BN}} & RCA    & \textbf{0.0000} & 0.2113 & \textbf{0.0002} & \textbf{0.0025} & 0.0404          & \textbf{0.0000} & \textbf{0.0094} & 0.1409 \\
                           & BCA    & \textbf{0.0000} & 0.0291 & \textbf{0.0021} & 0.0730          & \textbf{0.0022} & \textbf{0.0000} & \textbf{0.0013} & 0.0986
\\\hline
\end{tabular}
\end{table*}

\subsection{Effect of UR}

\begin{figure}[htpb]\centering
	\subfigure[]{\includegraphics[width=0.9\columnwidth,clip]{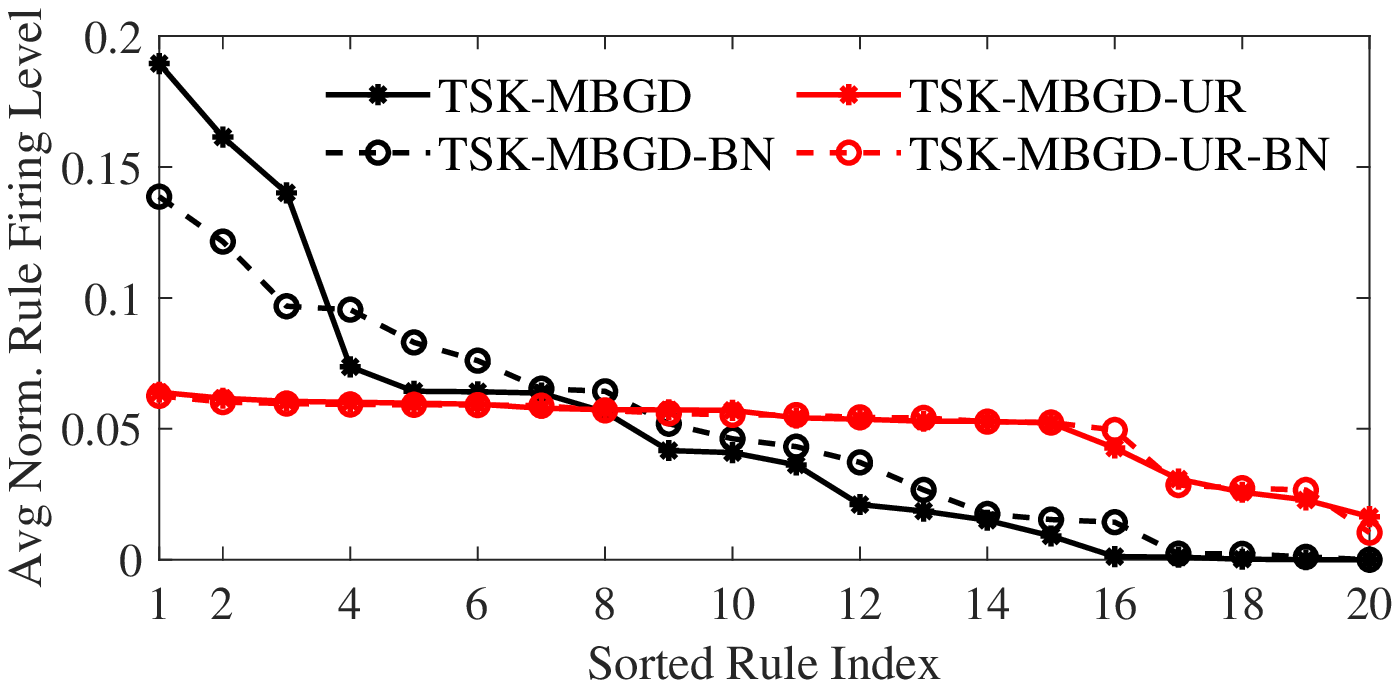}}
	\subfigure[]{\includegraphics[width=0.9\columnwidth,clip]{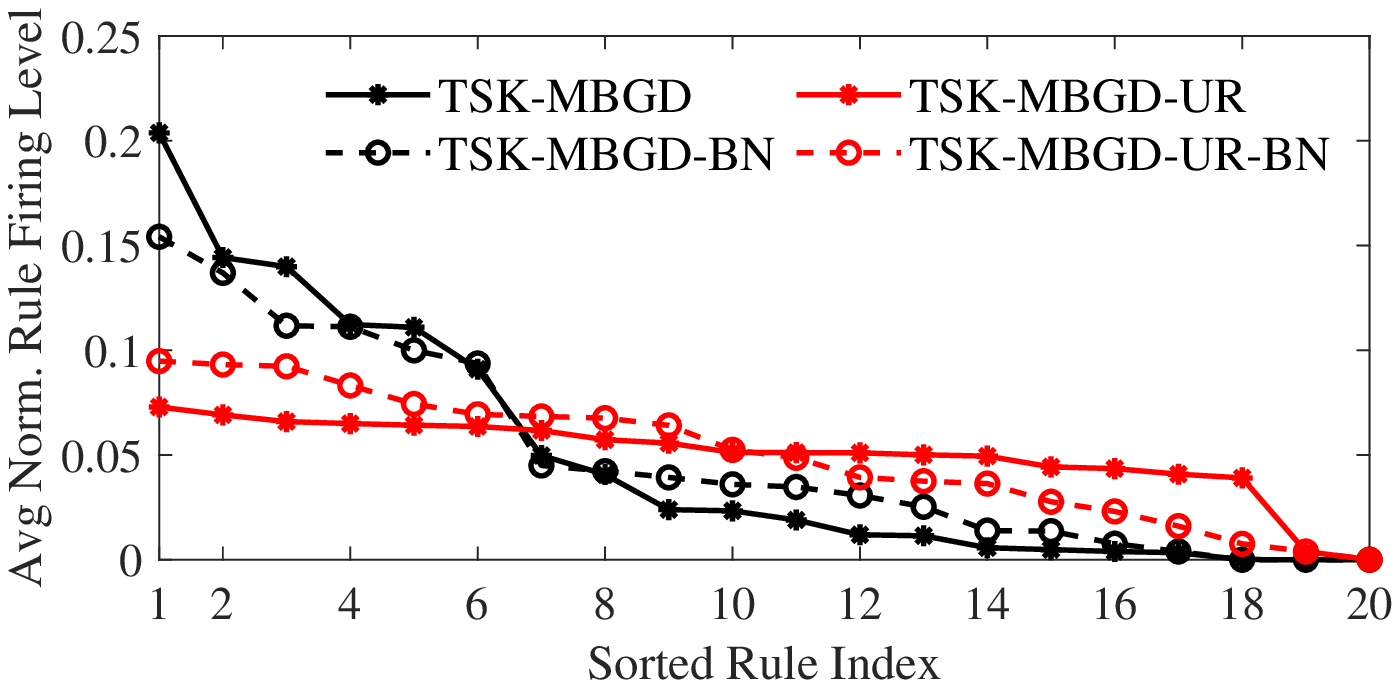}}
	\subfigure[]{\includegraphics[width=0.9\columnwidth,clip]{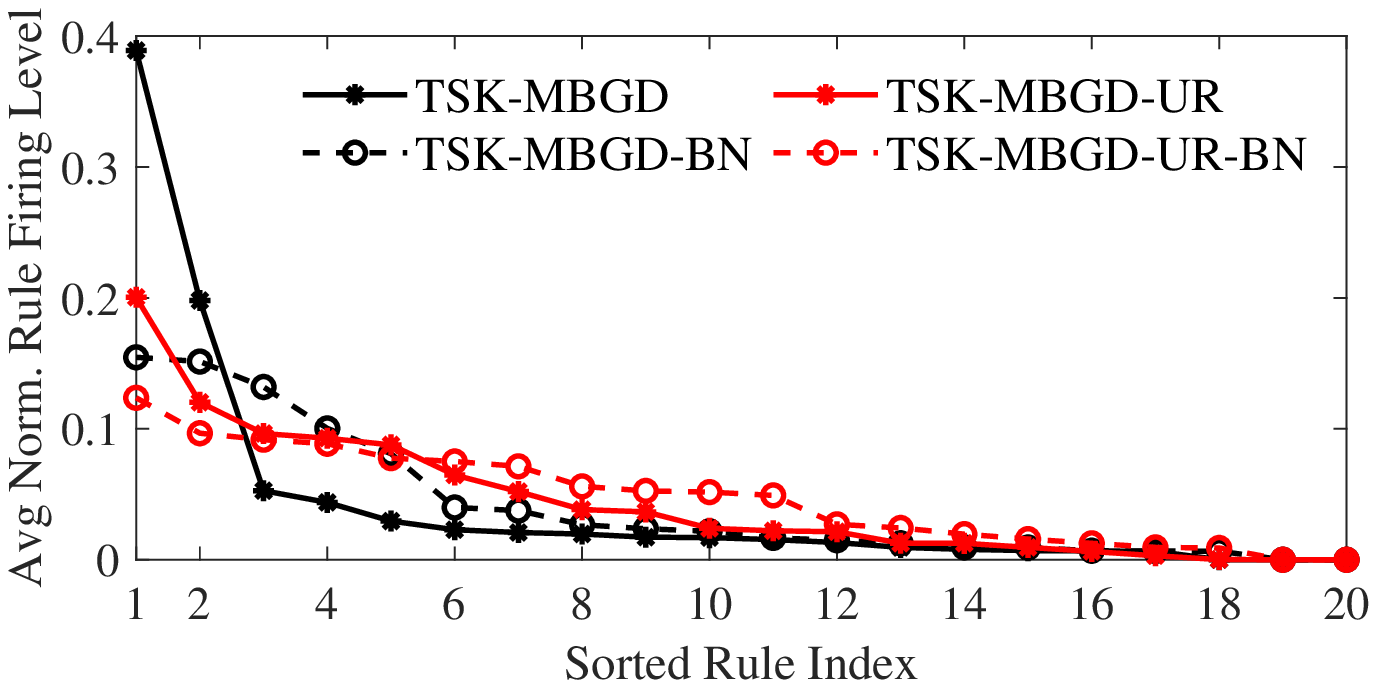}}
	\caption{Average normalized rule firing levels of \texttt{TSK-MBGD}, \texttt{TSK-MBGD-BN}, \texttt{TSK-MBGD-UR} and \texttt{TSK-MBGD-UR-BN} on (a) Satellite, (b) Vehicle, and (c) Biodeg datasets. } 	\label{fig:urfiring}
\end{figure}

As mentioned in Section~\ref{subsec:ur}, using MBGD to optimize the TSK fuzzy system may face the ``rich get richer" problem. To demonstrate this, Fig.~\ref{fig:urfiring} shows the average normalized firing levels of the rules on the entire dataset after the four MBGD-based TSK models were trained, on three representative datasets. For \texttt{TSK-MBGD}, a few ``richest" rules had much larger average firing levels than others, and hence the rules contributed significantly differently to the output. BN may help alleviate this problem a little bit, as the average normalized rule firing levels in \texttt{TSK-MBGD-BN} were more uniform than those in \texttt{TSK-MBGD}, which also resulted in better classification performances, as demonstrated in the previous subsection. However, UR had the most direct effect on alleviating the ``rich get richer" problem, as \texttt{TSK-MBGD-UR} (\texttt{TSK-MBGD-UR-BN}) had much more uniform average normalized rule firing levels than \texttt{TSK-MBGD} (\texttt{TSK-MBGD-BN}), and hence also better classification performance.

Note that we set $\tau=1/C$ in (\ref{eq:ur}), where $C=6$ for Satellite, $C=4$ for Vehicle, and $C=2$ for Biodeg. However, the actual average normalized rule firing levels were not exactly $\tau$ on these datasets. Our experiments showed that although UR cannot guarantee the average normalized rule firing levels to be around $\tau$, it can indeed make the rules fired more uniformly.

Why may making the rules fired more uniformly help improve the generalization performance? In \cite{wu2019functional} we pointed out that a TSK fuzzy system may be functionally equivalent to an adaptive stacking ensemble model, in which each rule can be viewed as a base learner, and the aggregation weights equal the corresponding rule firing levels. When the rule firing levels are more uniform, generally more rules are utilized in computing the output, i.e., more base learners are used in the stacking ensemble model, which may help improve the generalization performance.

To demonstrate this, we computed the entropy of the normalized rule firing levels for each input example:
\begin{align}
	E = -\sum_r^R \overline{f}_r~\log\overline{f}_r, 	\label{eq:entropy}
\end{align}
where $\overline{f}_r$ is the normalized firing level of the $r$-th rule. Generally, a larger entropy means more rules were fired.

\begin{figure}[htpb]\centering
	\subfigure[]{\includegraphics[width=0.95\columnwidth,clip]{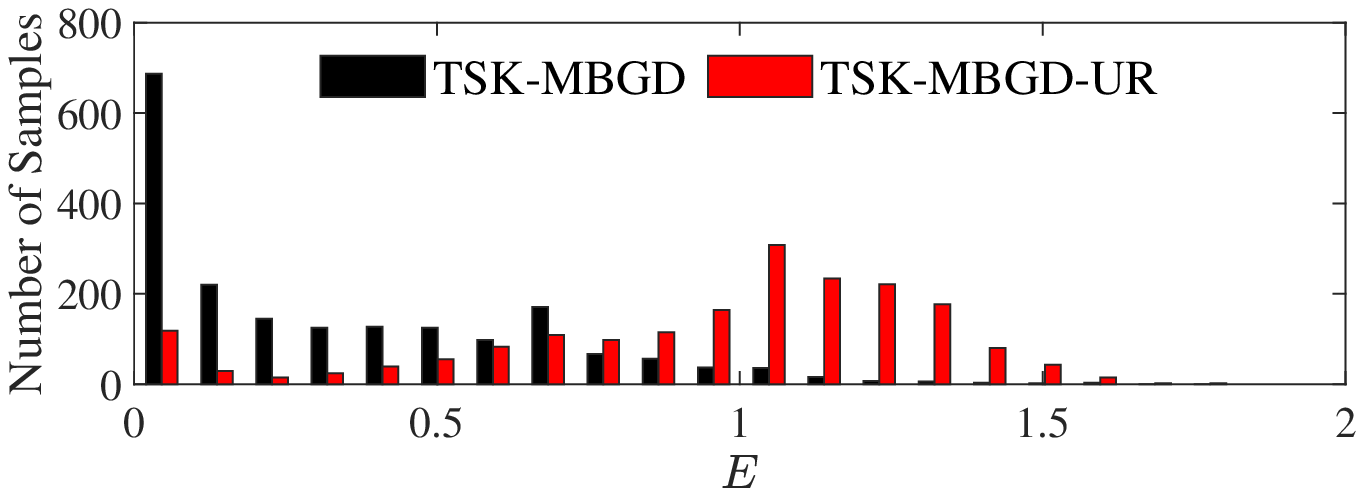}}
	\subfigure[]{\includegraphics[width=0.95\columnwidth,clip]{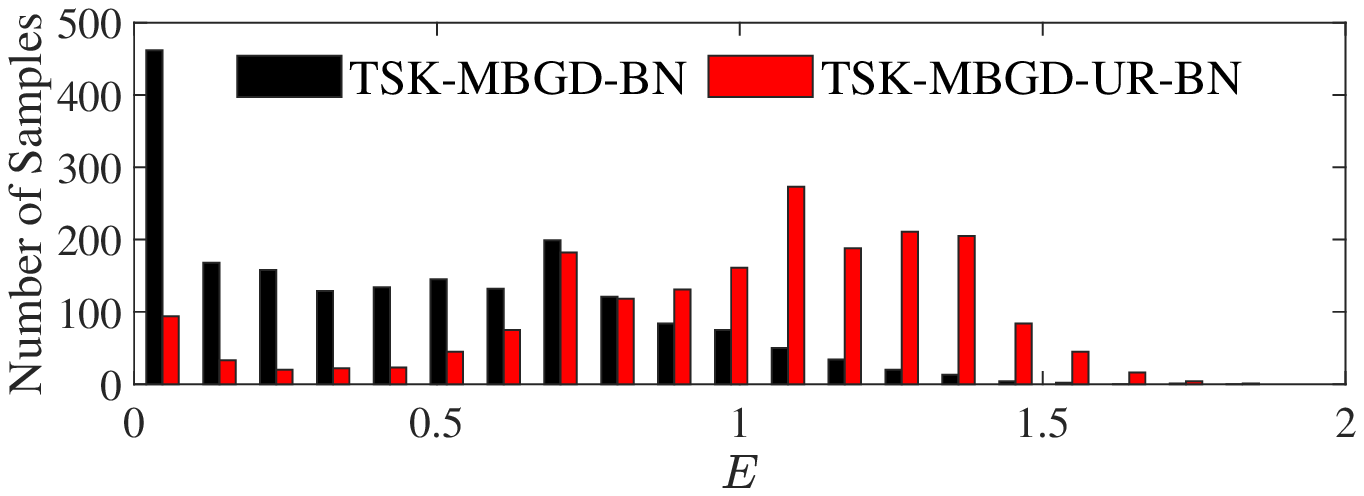}}
	\caption{Histogram of the normalized rule firing level entropy $E$ of (a) \texttt{TSK-MBGD} and \texttt{TSK-MBGD-UR}, and, (b) \texttt{TSK-MBGD-BN} and \texttt{TSK-MBGD-UR-BN}, on the Satellite dataset.} 	\label{fig:entro}
\end{figure}

Fig.~\ref{fig:entro} shows the histogram of the entropy distributions on the Satellite dataset. When training TSK fuzzy systems without UR, many samples had close to zero $E$, i.e., all except one rule had firing levels close to zero. When UR was added, the number of examples with close to zero $E$ decreased significantly, i.e., more rules with larger firing levels were used in computing the output.

\subsection{Effect of BN}\label{subsec:bnres}

We also used the Satellite dataset to analyze the effect of BN.

\begin{figure}[htpb]\centering
	\subfigure[]{\includegraphics[width=0.9\columnwidth,clip]{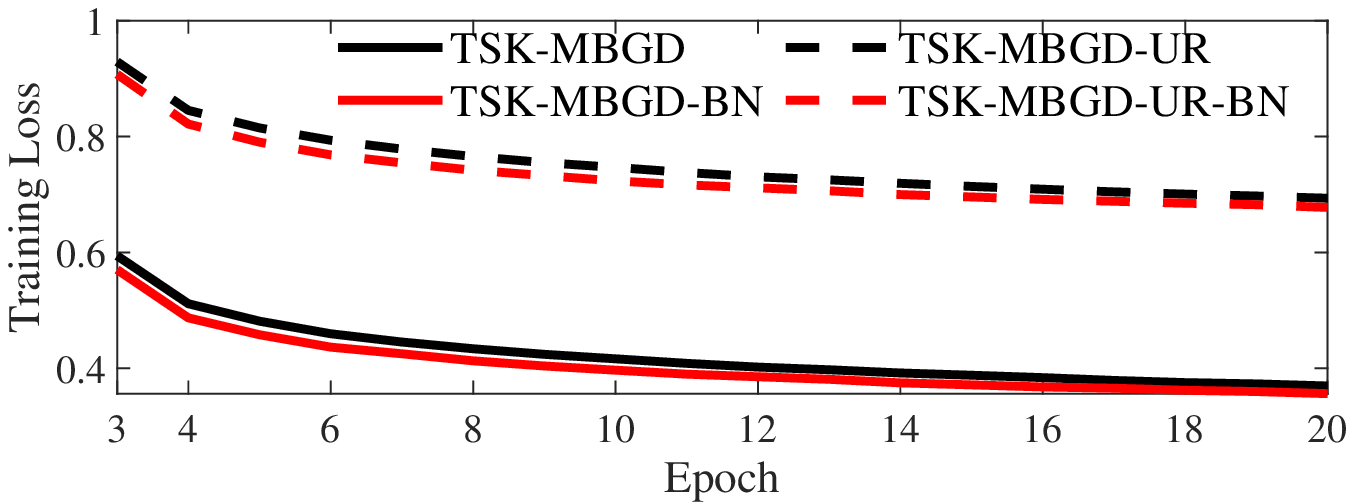}\label{subfig:histb}}
	\subfigure[]{\includegraphics[width=0.9\columnwidth,clip]{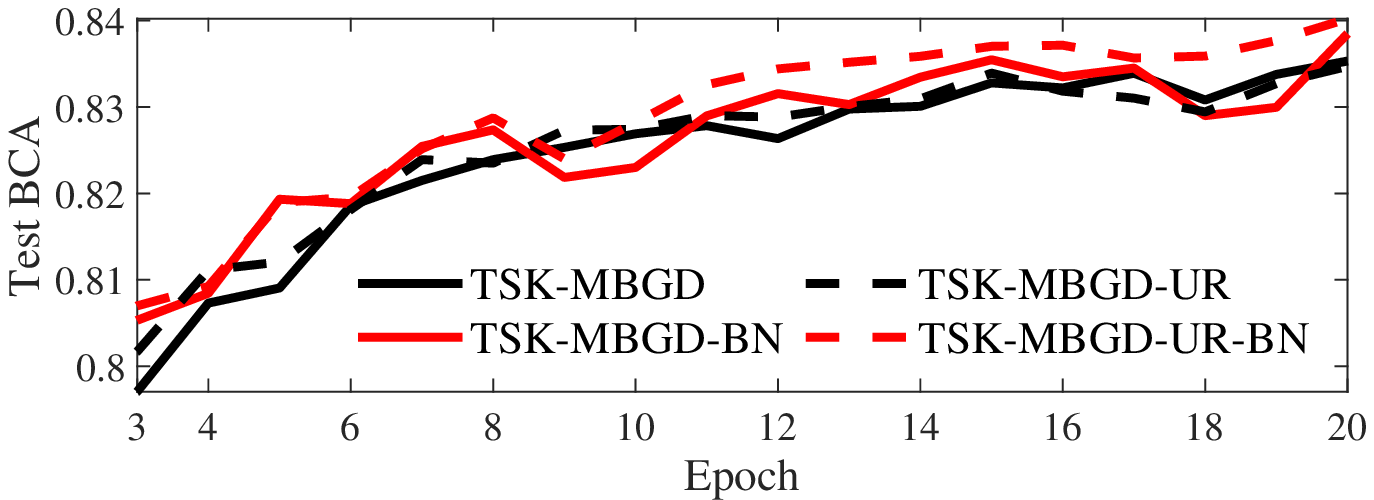}\label{subfig:hista}}
	\subfigure[]{\includegraphics[width=0.9\columnwidth,clip]{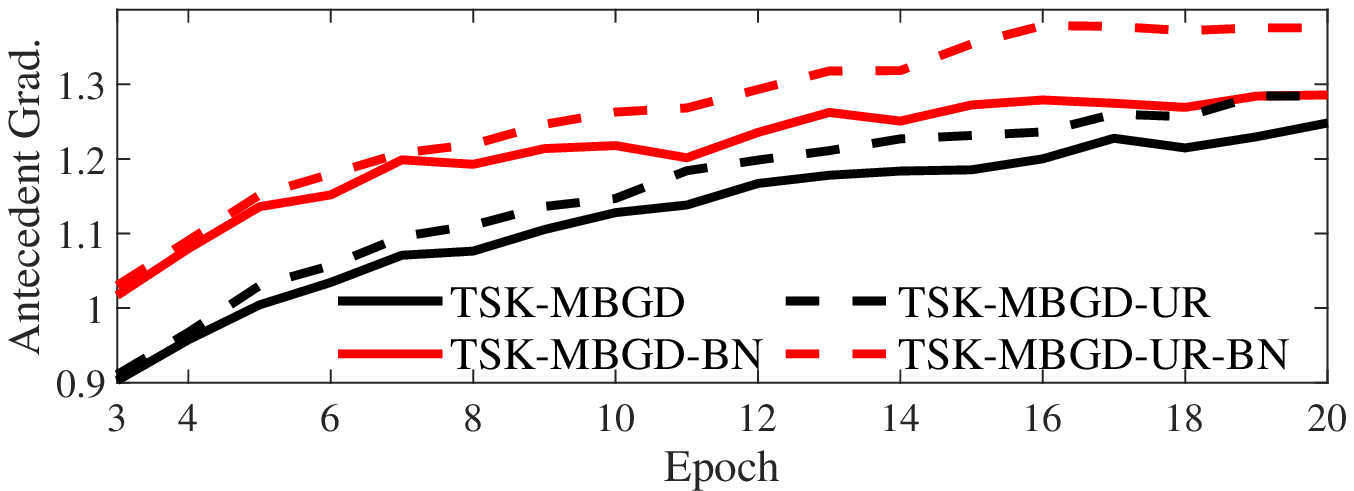}\label{subfig:grada}}
	\subfigure[]{\includegraphics[width=0.9\columnwidth,clip]{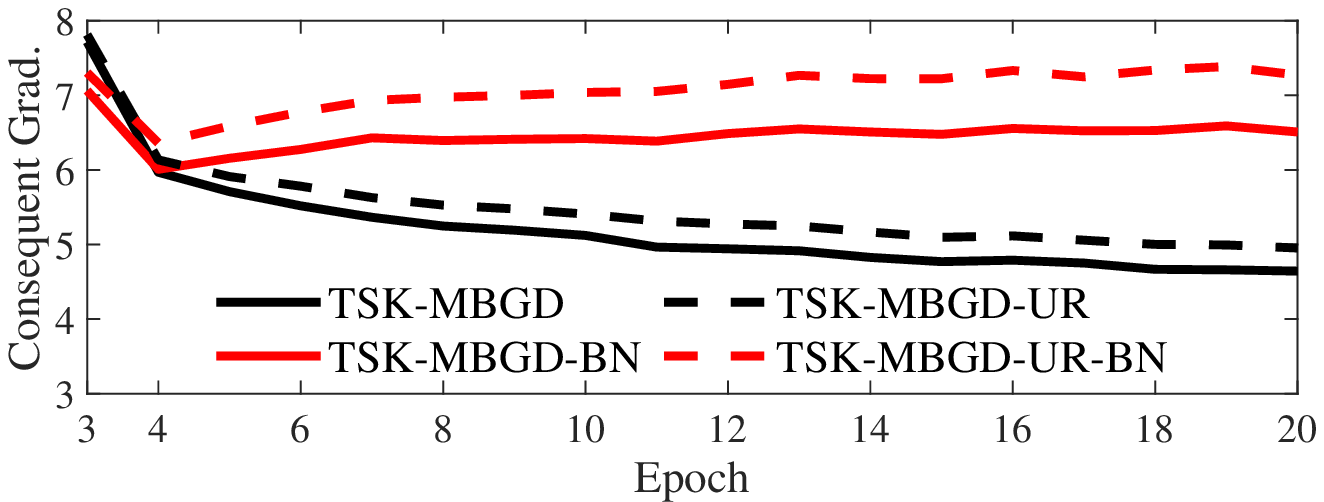}\label{subfig:gradb}}
		\caption{(a) Training loss, (b) test BCA, (c) L1 norm of the antecedent parameters' gradients, and (d) L1 norm of the consequent parameters' gradients, in the first 20 training epochs on the Satellite dataset. The horizontal axis starts from 3 epochs so that the differences among the curves can be more clearly visualized.} 	\label{fig:bn_hist}
\end{figure}

We set the UR weight $\lambda=1$ and recorded the training loss and test BCA in the first 20 training epochs. This process was repeated 10 times, and the average results are shown in Figs.~\ref{subfig:histb} and \ref{subfig:hista}, respectively. BN resulted in smaller training losses and better generalization performances in testing.

There is still no agreement on theoretically why BN is helpful in optimizing deep neural networks~\cite{santurkar2018does}; thus, it is also challenging to analyze theoretically why BN can help the optimization of TSK fuzzy systems. Nevertheless, we performed an empirical study to peek into this, by recording the L1 norm of the antecedent parameters' gradients and the L1 norm of the consequent parameters' gradients in the first 20 training epochs on the Satellite dataset. The results are shown in Figs.~\ref{subfig:grada} and \ref{subfig:gradb}, respectively. BN significantly increased the gradients of both antecedent and consequent parameters. With the same learning rate, this can expedite the convergence.

We also evaluated the performances of the two BN variants introduced in Section~\ref{subsec:bn}. The BCAs of \texttt{TSK-MBGD-UR}, \texttt{TSK-MBGD-UR-BN}, \texttt{TSK-MBGD-UR-GBN} and \texttt{TSK-MBGD-UR-RBN} are shown in Table~\ref{tab:bcabnv}. \texttt{TSK-MBGD-UR-BN} performed the best, and \texttt{TSK-MBGD-UR-GBN} the worst. Since \texttt{TSK-MBGD-UR-RBN} had more parameters to optimize, its training was not as stable as \texttt{TSK-MBGD-UR-BN} and \texttt{TSK-MBGD-UR-GBN}. Therefore, \texttt{TSK-MBGD-UR-BN} is the best choice.

\begin{table}[htpb]\centering
\caption{Average BCAs of the three BN variants on the 12 datasets.} \label{tab:bcabnv}
\setlength{\tabcolsep}{1mm}
\begin{tabular}{ccccc}\hline
\multirow{2}{*}{Dataset}        & \texttt{TSK-MBGD}          & \texttt{TSK-MBGD}      & \texttt{TSK-MBGD}       & \texttt{TSK-MBGD} \\
& \texttt{-UR}          & \texttt{-UR-BN}      & \texttt{-UR-GBN}       & \texttt{-UR-RBN} \\\hline
Vehicle           & 0.7127          & \textbf{0.7930}  & 0.7261 & 0.7679              \\
Biodeg            & 0.8390          & 0.8439          & 0.8422 & \textbf{0.8440}      \\
DRD               & 0.6717          & \textbf{0.6729} & 0.6636 & 0.6650               \\
Yeast             & 0.4946          & 0.5332          & 0.4352 & \textbf{0.5339}      \\
Steel             & 0.7245          & \textbf{0.7515} & 0.7332  & 0.7219              \\
IS & 0.8559          & \textbf{0.9501} & 0.9115 & 0.8938               \\
Abalone           & 0.4791          & \textbf{0.5402} & 0.4924 & 0.5275               \\
Waveform21        & \textbf{0.8362} & 0.8233          & 0.8232 & 0.8334               \\
Page-blocks       & 0.6033          & \textbf{0.6710} & 0.5912  & 0.6333              \\
Satellite         & 0.8679          & \textbf{0.8700} & 0.8679  & 0.8216              \\
Clave             & 0.6374          & 0.6421          & 0.6090   & \textbf{0.6442}    \\
MAGIC             & 0.8225          & 0.7934    & \textbf{0.8319}       & 0.8318     \\\hline
Average & 0.7121          &  \textbf{0.7404}       & 0.7106  & 0.7265  \\\hline
\end{tabular}
\end{table}

\subsection{Effect of the Batch Size}

The batch size is an important hyper-parameter in MBGD-based optimization. It determines the memory requirement and the convergence speed in training. A larger batch size leads to faster convergence but also requires more memory. In \cite{keskar2016large}, the authors analyzed the effect of the batch size on the generalization performance. Their results showed that using a larger batch size causes degradation in the model generalization performance, because it tends to converge to a shaper minimum, which makes the model sensitive to noise. A similar finding was presented in \cite{masters2018revisiting} that a smaller batch size leads to more stable and reliable training. However, since we used the mean, standard deviation and mean firing level of each batch to compute the losses, too small batch size may also lead to poor performance.

\begin{figure}[htpb]\centering
\includegraphics[width=\columnwidth, clip]{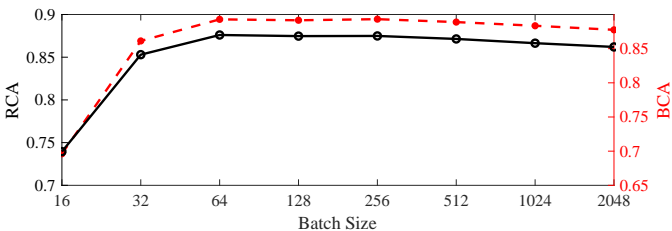}
\caption{Average RCAs and BCAs of \texttt{TSK-MBGD-UR-BN} on the Satellite dataset, using different batch sizes.} \label{fig:batch}
\end{figure}

We validated our model on the Satellite dataset with batch size varying from 16 to 2,048. The test RCAs and BCAs averaged over 30 runs are shown in Fig.~\ref{fig:batch}. The test performance decreased with too small or too large batch sizes. For \texttt{TSK-MBGD-UR-BN}, it seems that a batch size within [64, 256] is a good choice.

\section{Conclusions and Future Research}\label{sec:conclusion}

TSK fuzzy systems are powerful and frequently used machine learning models, for both regression and classification. However, they may not be easily applicable to large and/or high-dimensional datasets. Our very recent research \cite{drwuGD2019} proposed an MBGD-based efficient and effective training algorithm (MBGD-RDA) for TSK fuzzy systems for regression problems. This paper has proposed an MBGD-based algorithm, \texttt{TSK-MBGD-UR-BN}, to train TSK fuzzy systems for classification problems. It can deal with both small and big data with different dimensionalities, and may be the only algorithm that can train a TSK fuzzy classifier on big and high-dimensional datasets. \texttt{TSK-MBGD-UR-BN} integrates two novel techniques, which are also first proposed in this paper:
\begin{enumerate}
\item UR, which is a regularization term in the loss function to ensure that all rules are fired similarly on average, and hence to improve the generalization performance.
\item BN, which normalizes the inputs in computing the rule consequents to speedup the convergence and to improve the generalization.
\end{enumerate}

Experiments on 12 UCI datasets from various domains, with varying size and feature dimensionality, demonstrated that each of UR and BN has its own unique advantages, and integrating them can achieve the best classification performance. \texttt{TSK-MBGD-UR-BN}, together with MBGD-RDA proposed in \cite{drwuGD2019}, shall greatly promote the applications of TSK fuzzy systems in both classification and regression, especially for big data problems.

The proposed \texttt{TSK-MBGD-UR-BN} also has some limitations, which will be addressed in our future research. First, for very high dimensional data, fuzzy partitions of the input space become very complicated, and numeric underflow may happen when the product $t$-norm is used. Further research shall consider rules that automatically select the most relevant attributes as the antecedents. Second, we shall investigate how to improve the interpretability of data-driven TSK fuzzy systems. This is also partially linked to the first problem, as reducing the number of antecedents can improve the interpretability of the rules.


\end{document}